\def \JohnVersion{
  \documentclass[letterpaper, 12 pt, conference]{IEEEtran}
  \usepackage{setspace}
  \doublespacing
  \onecolumn
  \setstretch{2}
}

\documentclass[journal]{IEEEtran}

\usepackage[backend=biber,style=ieee]{biblatex}
\usepackage{common}
\usepackage{this_doc}
\usepackage{figures}
\usepackage{tables}
\usepackage{algorithms}
\usepackage{problems}
\usepackage{tikz}
\usetikzlibrary{matrix}
\usepackage{nicematrix}
\usepackage{balance}
\usepackage{flushend}

\usepackage[en-US]{datetime2}
\newcommand{\mydate}[1]{%
  \DTMsavetimestamp{creation}{#1T00:00:00-05:00}%
  \date{\DTMusedate{creation}}%
  {\DTMsetstyle{pdf}%
  \pdfinfo{
    /CreationDate (\DTMuse{creation})
  }}%
}
\mydate{2022-1-28}
\title{Certifiably Correct Range-Aided SLAM}

\author{
  Alan Papalia, Andrew Fishberg, Brendan W. O'Neill, Jonathan P. How, \\
   David M. Rosen, John J. Leonard
\thanks{
 This work was supported by the MIT Lincoln Laboratory Autonomous Systems Line
 which is funded by the Under Secretary of Defense for Research and Engineering
 through Air Force Contract No. FA8702-15-D-0001, ONR grants N00014-18-1-2832,
 N00014-23-12164, and N00014-19-1-2571 (Neuroautonomy MURI), the MIT Portugal
 Program, the National Defense Science and Engineering Graduate fellowship, the
 Tillman Scholar program, and the Department of Energy/National Nuclear Security
 Administration under Award Number(s) DE-NA0003921.
}
\thanks{ Alan Papalia, Andrew Fishberg, Brendan O'Neill, Jonathan P. How, and
John J. Leonard are with the \MIT, \MITaddr, (e-mail: apapalia@mit.edu; fishberg@mit.edu; oneillb@mit.edu; jhow@mit.edu;
jleonard@mit.edu) Alan Papalia is also with the Woods Hole Oceanographic
Institution, Woods Hole, MA 02543, USA.}
\thanks{David M. Rosen is with Northeastern University,
Boston, MA 02115, USA, ({email: d.rosen@northeastern.edu})}
}

\begin{document}


\maketitle

\begin{tikzpicture}[overlay, remember picture]
\path (current page.north east) ++(-4.4,-0.4) node[below left] {
This paper has been accepted for publication in the IEEE Transactions on Robotics.
};
\end{tikzpicture}
\begin{tikzpicture}[overlay, remember picture]
\path (current page.north east) ++(-6.7,-0.8) node[below left] {
Please cite the paper as:
A. Papalia,
A. Fishberg,
};
\end{tikzpicture}
\begin{tikzpicture}[overlay, remember picture]
\path (current page.north east) ++(-6.5,-1.2) node[below left] {
B. O'Neill,
J. P. How,
D. M. Rosen,
J. J. Leonard,
};
\end{tikzpicture}
\begin{tikzpicture}[overlay, remember picture]
\path (current page.north east) ++(-3.6,-1.6) node[below left] {
``Certifiably Correct Range-Aided SLAM'', \emph{IEEE Transactions on Robotics (T-RO)}, 2024.
};
\end{tikzpicture}

\begin{abstract}
    We present the first algorithm to efficiently compute certifiably optimal solutions to range-aided simultaneous localization and mapping (RA-SLAM) problems. Robotic navigation systems increasingly incorporate point-to-point ranging sensors, leading to state estimation problems in the form of RA-SLAM.  However, the RA-SLAM problem is significantly more difficult to solve than traditional pose-graph SLAM: ranging sensor models introduce non-convexity and single range measurements do not uniquely determine the transform between the involved sensors. As a result, RA-SLAM inference is sensitive to initial estimates yet lacks reliable initialization techniques.  Our approach, certifiably correct RA-SLAM (CORA), leverages a novel quadratically constrained quadratic programming (QCQP) formulation of RA-SLAM to relax the RA-SLAM problem to a semidefinite program (SDP). CORA solves the SDP efficiently using the Riemannian Staircase methodology; the SDP solution provides both (i) a lower bound on the RA-SLAM problem's optimal value, and (ii) an approximate solution of the RA-SLAM problem, which can be subsequently refined using local optimization. CORA applies to problems with arbitrary pose-pose, pose-landmark, and ranging measurements and, due to using convex relaxation, is insensitive to initialization.  We evaluate CORA on several real-world problems. In contrast to state-of-the-art approaches, CORA is able to obtain high-quality solutions on all problems despite being initialized with random values.  Additionally, we study the tightness of the SDP relaxation with respect to important problem parameters: the number of (i) robots, (ii) landmarks, and (iii) range measurements. These experiments demonstrate that the SDP relaxation is often tight and reveal relationships between graph connectivity and the tightness of the SDP relaxation.
\end{abstract}
\vspace*{-3mm}

\IEEEpeerreviewmaketitle

\section{Introduction}
\label{sec:introduction}

Reliable range-aided simultaneous localization and mapping (RA-SLAM) algorithms
are critical to enabling autonomous navigation in a variety of environments.
RA-SLAM application domains span underwater \cite{pelletier22ral,Paull14joe}, in
air \cite{guo2017ijmav}, underground \cite{funabiki21ral}, and planetary
\cite{boroson20iros} environments. RA-SLAM,
also known as range-only SLAM \cite{blanco08icra,boots13icml},
    extends pose-graph SLAM
    \cite{Cadena16tro,rosen21annualreviews} by allowing for arbitrary combinations
    of relative pose measurements and point-to-point range measurements between
    robot pose and landmark variables.
Range sensing possesses the
notable advantage that measurements often have known data-association, a primary failure
mode of modern navigation systems \cite{rosen21annualreviews}. Additionally, in
many environments (e.g., underwater) range measurements are critical to
long-duration state estimation. Despite range sensing's advantages and broad
applicability, RA-SLAM approaches lack the optimality guarantees and robustness
to initialization found in modern pose-graph SLAM \cite{rosen19ijrr}.

\TitleFigure

The difficulties surrounding RA-SLAM arise from the structure of the
\textit{maximum a posteriori} (MAP) estimation problem, which leads to a
nonlinear least-squares (NLS) problem \cite{Dellaert06ijrr,Kaess08tro}. As the
NLS formulation is non-convex, standard approaches only guarantee locally
optimal solutions and cannot distinguish local from global optimality. While
these issues have been circumvented in pose-graph SLAM \cite{rosen19ijrr}, range
measurements take a fundamentally different form than pose measurements in the
MAP problem, which prevents application of existing techniques.

This work presents CORA, an algorithm capable of returning certifiably optimal
estimates to the RA-SLAM problem (see \cref{fig:title-figure} for overview).
CORA solves an SDP relaxation of the RA-SLAM
problem to obtain (i) a lower bound on the optimal value of and (ii) an
approximate solution to the original RA-SLAM problem. This is made capable by a
novel reformulation of the RA-SLAM problem into a quadratically constrained
quadratic program (QCQP), which then leads to a natural SDP relaxation
\cite{shor1987quadratic}. We demonstrate how to apply the Riemannian Staircase
approach \cite{boumal15arxiv,boumal16neurips} to obtain a solution to the SDP
relaxation. The resulting SDP solution is projected to the feasible set of the
original RA-SLAM problem and locally optimized to produce a final estimate. The
difference between the cost of this final estimate and the SDP lower bound
provides an upper bound on the solution suboptimality.

We evaluate CORA's performance on a range
of real-world RA-SLAM datasets, including a 3D aerial drone data set collected
as part of this work. We show that, unlike existing approaches which often
perform well but are susceptible to poor initialization and local optima, CORA
consistently obtains high-accuracy solutions regardless of initialization.
Additionally, we characterize the quality of the SDP lower bound and perform
parameter sweeps to observe how common parameters (e.g., number of distinct
robots) in RA-SLAM problems affect the tightness (the minimum attainable
suboptimality gap) of the proposed SDP relaxation. While the relaxation's
tightness may not affect the quality of the final estimate, it affects the
ability to obtain a certifiably optimal solution to the unrelaxed problem.
This parametric study provides insight into the structure of
RA-SLAM problems and suggests which characteristics lead to more difficult
optimization problems.

We summarize the contributions of this paper as follows:
\begin{itemize}
    \item The first RA-SLAM algorithm capable of producing certifiably optimal
          estimates.
    \item A parametric study of tightness of the proposed SDP relaxation.
    \item A performant C++ implementation, including all experimental data
          presented in this paper\footnote{\url{\RepoURL}}.
\end{itemize}

\section{Related Works}
\label{sec:related-works}
\ProbOverviewFigure

We discuss existing RA-SLAM approaches as well as
works in certifiable estimation.
This paper represents a methodological contribution towards the inference
subproblem in RA-SLAM. Our discussion of RA-SLAM works will briefly
mention applications of RA-SLAM, but will primarily focus on inference
methodologies for the RA-SLAM problem. We will not discuss many works (e.g.,
\cite{queralta2022vio,funabiki21ral,nguyen2022viral,fishberg2022multi}) which
address other important aspects of RA-SLAM, such as measurement extraction or
outlier rejection.

Furthermore, discussion of certifiable estimation will separate works by
estimation methodology. Certification in robotic perception typically describes
a certificate of optimality of some estimated quantity, where the estimation
procedure is defined as an optimization problem
\cite{barfoot2017state,bandeira16probablycertifiably,rosen20wafr}. Works in
certifiable estimation generally are either (i) an optimizer paired with a
global optimality certification scheme or (ii) an exhaustive search approach. A
notable subset of the former class of works falls under \textit{certifiably
correct} methods, which can efficiently recover \textit{provably} optimal
solutions to a problem within a given problem regime
\cite{bandeira16probablycertifiably}.

\subsection{Range-Aided SLAM}

Early works in RA-SLAM considered only odometry and point-to-point range
measurements from robot poses to landmarks \cite{djugash14ijfr}. This
formulation leads to simple, computationally
efficient systems: standard odometry and range sensors can be
used, data association is automatically known, and the number of landmarks is
equal to the number of ranging beacons. More recent works have
additionally incorporated visual landmarks \cite{nguyen2023vr,boroson20iros}
and explored inter-robot ranging \cite{boroson20iros,shule2020uwb}.

In the inference stage, these systems still battled the unique
nonlinearities introduced by range measurements. To address these
nonlinearities, previous range-only SLAM works developed approaches utilizing:
standard extended Kalman filters (EKFs) \cite{Newman03icra,menegatti09icra},
polar-coordinate EKFs \cite{Djugash09iros,djugash09springer}, particle filters
\cite{gonzalez09ras,blanco08icra,blanco08iros}, and nonlinear least-squares
optimization \cite{herranz14icra,cao2021vir}. These approaches all demonstrated
success, particularly when accurately initialized. However, unlike CORA, these
approaches are largely contingent on a quality initialization and provide no
guarantees (\textit{a priori} or \textit{a posteriori}) on the quality of the
returned estimate.

\subsection{Sufficient Conditions for Global Optimality}

Many certification schemes leverage relationships between quadratically
constrained quadratic programs (QCQPs) and SDPs to construct certifiers based on
Lagrangian duality. These certifiers establish sufficient conditions for
optimality which, if satisfied, guarantee global optimality
\cite{jeyakumar07mathprog,li12optimizationtheory,rosen20wafr}.

Additionally, works in certification can be separated by the estimation
methodology proposed; works typically pair a certifier with a local-optimizer or
a SDP relaxation. The SDP approaches solve a relaxation of the original problem
and then project the relaxed solution to the feasible set of the original
problem. Notable SDP-based works
\cite{rosen19ijrr,tian21tro} established conditions when strong optimality
guarantees exist.  Importantly, unlike local optimization, the estimates from
SDP methods are
initialization-independent.

\subsubsection*{Local-optimization and Verification}

Karush-Kuhn-Tucker (KKT) based sufficient conditions were derived from various
QCQPs and paired with
local optimizers to solve problems in point-cloud registration
\cite{iglesias20cvpr,yang20tro}, multi-view geometry
\cite{garcia-salguero21ivc}, pose-graph SLAM
\cite{briales16iros,carlone15iros}, and range-only localization
\cite{dumbgen22ral}. In addition to KKT-based certificates, a number of
computer vision works developed certification schemes based on
proving problem convexity over the feasible set
\cite{garcia-salguero21siamjis,hartley13bijcv,olsson09cvpr}.

\subsubsection*{SDPs and Interior Point Methods}

Many approaches in robotic perception formulate problems as a QCQP,
solve an SDP relaxation of the QCQP \cite{bao11mathprog}, and then project the
SDP solution to the feasible set of the original problem. Often the SDP
relaxation is exact, and thus the projection is a certifiably optimal solution
to the original problem. Previous approaches applied standard SDP solvers to
problems in
geometric registration \cite{briales17cvpr,olsson08icpr,khoo16tip},
robust point-cloud registration \cite{yang19rss},
relative pose estimation \cite{briales18cvpr,garcia-salguero22jmiv},
triangulation \cite{aholt12lncs},
anonymous bearing-only multi-robot localization \cite{wang22arxiv},
essential matrix estimation \cite{zhao20tpami},
pose-graph SLAM \cite{carlone16tro},
and sensor calibration \cite{giamou19ral}.
While standard interior point methods scale poorly for large problems, these
approaches are successful on small problems and can use a broad set of solvers.

\subsubsection*{SDPs and the Riemannian Staircase}

SDP relaxations of robotic perception problems typically admit low-rank
solutions. A growing body of work applies the Riemannian Staircase methodology
\cite{boumal15arxiv,boumal16neurips} to leverage this low-rank structure and
more efficiently solve the SDP. Riemannian Staircase approaches differ from
interior point methods in that they (i) optimize over a
series of lower-dimension, non-convex, rank-restricted SDPs to solve the
original SDP and (ii) these methods leverage the fact that the feasible
sets of these rank-restricted problems take the form of simple manifolds,
enabling the use of Riemannian optimization methods to improve optimizer
performance. While the approach of solving a series of
rank-restricted SDPs can \textit{in principle} be applied to all SDPs
\cite{burer03mathprog,burer05mathprog,rosen20wafr}, usage of Riemannian
optimization depends on specific problem structure and does not apply to all
SDPs. Within robotics, the Riemannian Staircase has been applied to
pose-graph SLAM \cite{briales17ral,rosen19ijrr,fan20tro}, essential matrix
estimation \cite{karimian2023essential}, and range-only localization
\cite{halsted22arxiv}.

\subsubsection*{Placement of this Work}
This work falls under the Riemannian Staircase class of works. We present a SDP
formulation, certification procedure, Riemannian Staircase methodology, and
means for extracting an RA-SLAM estimate from the SDP solution. Our work
differs from recent, important papers in certification for range-only
localization \cite{dumbgen22ral,halsted22arxiv}, as a
novel problem formulation and certification methodology were necessary to
account for pose variables. Our certification approach builds upon the KKT-based
analyses of \cite{jeyakumar07mathprog,rosen20wafr}. Additionally, our problem
formulation and estimation procedure generalize key works in range-only
localization \cite{halsted22arxiv} and pose-graph SLAM
\cite{rosen19ijrr,briales17ral} to the case of RA-SLAM.

\subsection{Exhaustive Search}

Exhaustive search approaches largely use either polynomial root finding or
branch-and-bound (BnB) techniques to guarantee a globally optimal solution will
be found \cite{morrison2016branch,cox2013ideals}. Such \textit{a priori}
guarantees typically cannot be made for other approaches. These approaches come
at increased computational cost, particularly as problem sizes increase.

Polynomial solving via \Grobner~basis computation was applied to multi-view
geometry \cite{mirzaei11iccv,stewenius05iccv} and range-only
localization \cite{trawny10icra}. Similarly, \cite{mirzaei11icra} solved a
polynomial system via eigendecomposition to estimate relative camera pose.

BnB was applied to consensus maximization in rotation search
\cite{bazin12accv} and special Euclidean registration \cite{olsson09tpami}.
\cite{izatt20isrr} linearly approximated $\SOThree$ to solve pose
estimation as a mixed-integer convex program for improved BnB efficiency.
Similarly, \cite{yang16tpami} integrated the iterative closest point solver into
a BnB framework to efficiently perform point cloud registration. BnB was also
applied to the estimation of: camera focal length and relative
rotation \cite{bazin14eccv}, correspondence-free relative pose
\cite{campbell17iccv,fredriksson16cvpr}, essential matrices
\cite{hartley09ijcv,kneip13iccv}, and triangulation \cite{lu07lncs}.

\section{RA-SLAM as an SDP}
\label{sec:ra-slam-as-qcqp}

The RA-SLAM problem aims to estimate a collection of robot poses and landmark
locations from a set of relative pose and range measurements between these poses
and landmarks. In this section we demonstrate how the \textit{maximum a
        posteriori} (MAP) formulation of the RA-SLAM problem can be used to derive a
novel SDP. This SDP underpins many of the key properties of this work. We
outline the derivation of the SDP and the properties it provides in
\cref{fig:prob-overview}.

\subsection{MAP formulation of RA-SLAM}
\label{sec:ra-slam-map-formulation}

The MAP
formulation is based on generative measurement models with Langevin distributed
rotational noise \cite{chiuso08infosys} and Gaussian distributed translational
and ranging noise:
\begin{align}
        \label{eq:rot-noise-models}
        \nrot_{i j}  & =\trot_{i j} \prot_{i j},                           & \prot_{i j}  & \sim \langevin \left(I_{d}, \kappa_{i j}\right)
        \\
        \label{eq:tran-noise-models}
        \ntran_{i j} & =\ttran_{i j}+\ptran_{i j},                         & \ptran_{i j} & \sim \mathcal{N}\left(0, \tau_{i j}^{-1} I_{d}\right)
        \\
        \label{eq:dist-noise-models}
        \ndist_{i j} & =\lVert \ttran_i - \ttran_j \rVert_2 +\pdist_{i j}, & \pdist_{i j} & \sim \mathcal{N}\left(0, \sigma_{ij}^2\right)
\end{align}
where $\nrot_{ij}, \ntran_{ij}, \ndist_{ij}$ are noisy relative rotational,
translational, and range measurements. Similarly, $\trot_{ij}, \ttran_{ij},
        \tdist_{ij}$ are the true relative rotations, translations, and ranges. Finally,
$\prot_{ij}, \ptran_{ij}, \pdist_{ij}$ represent the noisy perturbations to the
measurements where the coefficients $\kappa_{ij}$, $\tau_{ij}$, and $(1 /
        (\sigma_{ij}^2))$ are, respectively, the rotational, translational, and ranging
measurement precisions.

From the measurement models of
\cref{eq:rot-noise-models,eq:tran-noise-models,eq:dist-noise-models} the MAP
formulation of RA-SLAM is as follows \cite{papalia23icra},
\footnote{
                Without loss of generality, this formulation doesn't distinguish
                landmark  variables from poses. This is to
                simplify the mathematical presentation but does not preclude use
                of landmarks or mean that `degenerate' poses are used. See
                Appendix \ref{sec:landmarks-in-problem-formulation} for further
                discussion.
}

\RaSlamMapProblem

\subsection{Obtaining the SDP Relaxation}

From \cref{prob:ra-slam-map} we derive a relaxed problem, which takes the form
of a QCQP and, in turn, leads to a convenient SDP relaxation. This QCQP
(\cref{prob:ra-slam-qcqp}) relaxes the special orthogonal constraint $\rot_i
        \in \SOd$ to an orthogonality constraint, $\rot_i \in \Orthogonald$. This
relaxation was found to have no significant impact on typical pose-graph SLAM solution
quality \cite{rosen19ijrr,briales17ral,tron15rssworkshop}.

Additionally, the range cost terms of \cref{prob:ra-slam-map} are modified. We
introduce auxiliary unit-norm vector variables, $\dist_{ij} \in \dvec$. This
reformulation of the range cost terms in \cref{prob:ra-slam-qcqp} is equivalent
to the terms presented in \cref{prob:ra-slam-map} in the sense that the optimal
cost and optimal translations $\tran_i$ are preserved for both problems
\cite[Lemma 1]{halsted22arxiv}. This reformulation is a critical step in our
approach, as it enables range measurements to be expressed in quadratic forms.

We present this relaxation in \cref{prob:ra-slam-qcqp}, noting that all costs
and constraints are now quadratic.

\RaSlamQcqpProblem

The advantage of \cref{prob:ra-slam-qcqp} over \cref{prob:ra-slam-map} is that
\cref{prob:ra-slam-qcqp} is a QCQP, and therefore admits a standard SDP
relaxation \cite{shor1987quadratic}. This SDP relaxation of
\cref{prob:ra-slam-qcqp} takes the form of,
\RaSlamSdpProblem

This SDP relaxation (\cref{prob:ra-slam-convex-sdp}) will form the basis of our
approach.

\section{Solving \texorpdfstring{\cref{prob:ra-slam-convex-sdp}}{Problem 3} via
  Riemannian Optimization}
\label{sec:solving-convex-sdp-via-riemannian-optimization}

While the SDP of \cref{prob:ra-slam-convex-sdp} is key to our approach, standard
interior-point SDP solvers cannot scale application-relevant sizes of
\cref{prob:ra-slam-convex-sdp} We derive a rank-restricted SDP from
\cref{prob:ra-slam-convex-sdp}, which substantially reduces problem
dimensionality and thus computational cost.  Furthermore, we demonstrate that
this specific rank-restricted SDP can be solved via Riemannian optimization,
lending improved computational efficiency and insights into the geometric
structure of the problem. We outline these derivations and their relationships
to \cref{prob:ra-slam-convex-sdp} in \cref{fig:prob-overview}.

\subsection{Rank-Restricted SDPs}
\label{sec:rank-restricted-sdp}

As we expect \cref{prob:ra-slam-convex-sdp} to have a low-rank solution, we
instead apply the Burer-Monteiro method \cite{burer03mathprog,burer05mathprog}
to pose the problem as a rank-restricted SDP. This rank-restricted SDP reduces
the problem dimensionality by substituting an assumed low-rank factorization for
the optimal solution $\matZ = \matXOuterProd, \matX \in \matXStateSpace$,
arriving at the following form:
\RaSlamRankRestrictedSdpProblem
In this rank-restricted SDP the problem variable $\matX$ is composed as
follows,
\StackedMatrixDefinition
where each element, e.g., $\rot_i \in \R^{\liftedDimension \times \dimension}$,
can be considered a natural lifting of the variables in the QCQP of
\cref{prob:ra-slam-qcqp} when $\liftedDimension >
        \dimension$.

For each $\rot_i \in \R^{\liftedDimension \times \dimension}$ variable there are
$\frac{d(d+1)}{2}$ orthonormality constraints in the form of $\tr (\matAi
        \rotOuterProd) = b_i$. E.g., for $\dimension = 2$ the constraint $\rotOuterProd
        = I_d$ can be defined by the following pairs ($\matA_{j \rot_i}, b_j$):
\begin{align*}
        \matA_{1 \rot_i} & = \begin{bmatrix}
                1 & 0 \\
                0 & 0 \\
        \end{bmatrix} & b_1 & = 1  \\
        \matA_{2 \rot_i} & = \begin{bmatrix}
                0   & 1/2 \\
                1/2 & 0   \\
        \end{bmatrix} & b_2 & = 0  \\
        \matA_{3 \rot_i} & = \begin{bmatrix}
                0 & 0 \\
                0 & 1 \\
        \end{bmatrix} & b_3 & = 1.
\end{align*}

Similarly, the unit-norm constraint for a single $\dist_{ij}$ can be expressed
as $\tr(\matA_{\dist_{ij}} \dist_{ij}^\top \dist_{ij}) = 1$, where
$\matA_{\dist_{ij}} = \begin{bmatrix} 1 \end{bmatrix}$.

As these constraints all act on a single variable each, to embed them into the
full state space $\R^{\sizeQcqpStateSpace \times \sizeQcqpStateSpace}$ they are
simply placed at the block-diagonal location corresponding to the variable they
constrain. We outline these constraints in more detail in Appendix
\ref{sec:constraint-matrix-definition}.

In the Burer-Monteiro factorization, $\liftedDimension$ constrains
the rank of the solution to \cref{prob:ra-slam-rr-sdp}, as $\rank(\matX) =
        \rank(\matXOuterProd) \leq \liftedDimension$. Indeed, the rank-restricted SDP of
\cref{prob:ra-slam-rr-sdp} is equivalent to the QCQP of \cref{prob:ra-slam-qcqp}
when $\liftedDimension = \dimension$. It follows that incrementing
$\liftedDimension$ provides an interpretable means of relaxing
\cref{prob:ra-slam-rr-sdp} by increasing the allowable rank of the solution.

This relaxation methodology is key to our approach. We often need to relax the
problem to escape local minima of the rank-restricted SDP but we can typically
find solutions to the SDP of \cref{prob:ra-slam-convex-sdp} at relaxations $p
        \ll \sizeQcqpStateSpace$.

\subsection{\texorpdfstring{\cref{prob:ra-slam-rr-sdp}}{Problem 4} as Riemannian Optimization}
\label{sec:rr-sdp-as-riemannian-optimization}

The feasible set of the rank-restricted SDP in
\cref{prob:ra-slam-rr-sdp} can be described as a product of Riemannian
manifolds.

We have established that the constraints of \cref{prob:ra-slam-rr-sdp} are
equivalent to $\rotOuterProd = I_d$ for all $\rot_i \in \R^{\liftedDimension
                \times \dimension}$ and $\lVert \dist_{ij} \rVert_2^2 = 1$ for all $\dist_{ij}
        \in \R^{\liftedDimension}$. The orthonormal constraint, $\rotOuterProd = I_d$ is
equivalent to the Stiefel manifold $\St (\liftedDimension, \dimension)$
\cite{rosen19ijrr}. Similarly, the unit-norm constraints, $\lVert \dist_{ij}
        \rVert_2^2 = 1$, are equivalent to $S^{\liftedDimension-1}$, the unit-sphere in $\R^\liftedDimension$.
Thus the feasible set of \cref{prob:ra-slam-rr-sdp} can be expressed as the
product of Riemannian manifolds for any $\liftedDimension$.

Recognizing this Riemannian structure to our problem (i) allows us to apply
mature Riemannian optimization techniques to solve the rank-restricted SDPs and
(ii) provides insights into the smoothness of the underlying constraints which
will prove useful in our certification methodology. We describe the
corresponding Riemannian optimization problem in
\cref{prob:ra-slam-riemannian-staircase}.

We emphasize that \cref{prob:ra-slam-riemannian-staircase} is exactly the
rank-restricted SDP of \cref{prob:ra-slam-rr-sdp}, reformatted to explicitly
state that the feasible set of \cref{prob:ra-slam-rr-sdp} is a product of
Riemannian manifolds.

\RaSlamRiemannianStairProblem

\section{Certifying Rank-restricted SDP Solutions}
\label{sec:optimality-certificates}

In this section we demonstrate how previous results in certifiable estimation
\cite{rosen20wafr} can be applied to the rank-restricted SDP
(\cref{prob:ra-slam-rr-sdp}) to determine whether a candidate solution
$\candidateSolution$ is globally optimal.

This globally optimality test centers around the \textit{certificate matrix}
$\certMat$, which is formed from the data matrix $\matQ$, the constraint
matrices $\matAi$, and their corresponding Lagrange
multipliers $\lambda_i$ at a KKT point:
\begin{equation}
    \label{eq:cert-matrix}
    \certMat \triangleq
    \matQ +
    \sumOverConstraints \lambda_\indexVar \matAi.
\end{equation}
If, given a KKT point $\candidateSolution$ of \cref{prob:ra-slam-rr-sdp}, we
obtain an $\certMat$ which is positive semidefinite, then $\candidateSolution$
solves both \cref{prob:ra-slam-convex-sdp} and \cref{prob:ra-slam-rr-sdp}. This
is because if $\certMat \isPSD$ then $\matZ = \matXOuterProd$ is a KKT point of
the SDP of \cref{prob:ra-slam-convex-sdp} and thus must be optimal.  Critically,
our approach to obtaining $\certMat$ requires that the linear independence
constraint qualification (LICQ) is satisfied. In the following subsections we
discuss the LICQ, prove it is satisfied for our problem, and then describe our
algorithm for performing certification.

\subsection{The Linear Independence Constraint Qualification}
\label{sec:licq}

The certificate matrix we devise is a function of the Lagrange multipliers
$\lambda$ at a given solution point $\candidateSolution$. However, to guarantee
a unique set of Lagrange multipliers the LICQ must be satisfied
\cite{wachsmuth13licq}. Without a unique set of Lagrange multipliers, the
optimality condition becomes a test for whether there exist \textit{some} valid Lagrange
multipliers which generate a $\certMat \isPSD$. This thus becomes a SDP feasibility
problem of the same dimension as our original SDP (\cref{prob:ra-slam-convex-sdp}) and
thus is computationally intractable for desired RA-SLAM problem sizes.

The LICQ requires that the gradients of the constraints are linearly
independent. For \cref{prob:ra-slam-rr-sdp}, the LICQ is equivalent to linear
independence of $\{ \nabla_{\matX} \StackedMatrixConstraintLhs, ~~ i=1, \ldots,
    \numConstraints \}$ evaluated at $\matX = \candidateSolution$.  This condition
is equivalent to the following \textit{adjoint constraint Jacobian},
\begin{equation}
    \label{eq:constraint-jacobian}
    \ConstraintJacobian \in \ConstraintJacobianSize
    \triangleq \ConstraintJacobianDefinition
\end{equation}
having full column rank, where $\vect (\cdot)$ is the row-wise vectorization
operator.

\subsection{\texorpdfstring{\cref{prob:ra-slam-rr-sdp}}{Problem 4} satisfies the LICQ}
\label{sec:licq-proof}

We prove that the LICQ is satisfied for all feasible points of
\cref{prob:ra-slam-rr-sdp}. Our proof utilizes the fact that the constraints of
\cref{prob:ra-slam-rr-sdp} are defining functions of the embedded submanifolds:
$\St (\liftedDimension, \dimension)$ and $S^{\liftedDimension-1}$ \cite[Chapter
7]{boumal2023introduction}.

\begin{theorem}
    \label{thm:licq}
    The LICQ is satisfied for all feasible points of \cref{prob:ra-slam-rr-sdp}.
\end{theorem}

\textit{Proof:}
Recall that the constraints of \cref{prob:ra-slam-rr-sdp} have block-diagonal
sparsity patterns. Each constraint only accesses a single variable, $\rot_i$ or
$\dist_{ij}$. As a result, there should be an ordering of constraints and
variables which results in a block-diagonal adjoint constraint Jacobian.

Without loss of generality, we define an ordering over the constraints.
First group the constraints by their associated variable. We define
$\{\matA_{\rot_i}\}$ as the $\frac{d(d+1)}{2}$ constraints associated with
$\rot_i$ and $\{\matA_{\dist_{ij}}\}$ as the single constraint associated with
$\dist_{ij}$. The ordering within each group is arbitrary. Furthermore, order
these groups by the order of the associated variables as seen in
\cref{eq:stacked-matx}.
\comment{
With $A \prec B$ denoting that $A$ precedes $B$,
the ordering should be:
\begin{equation}
    \{\matA_{\rot_i}\} \prec \{\matA_{\rot_{j}}\}
    \prec \{\matA_{\dist_{a}}\} \prec \{\matA_{\dist_{b}}\}
\end{equation}
for any $i < j$ and $a < b$.
}

Given this constraint ordering and the variable ordering of
\cref{eq:stacked-matx}, the adjoint constraint Jacobian of \cref{eq:constraint-jacobian}
takes a block-diagonal form.
\comment{
\begin{equation}
    \ConstraintJacobian \triangleq
    \begin{bNiceMatrix}
        \mathrm{X}  &   &   & \\
        & \Ddots  &   &  \\
        &  &   &  \\
        &  &   & \mathrm{X} \\
        & & & \\
    \end{bNiceMatrix}.
\end{equation}
}
The first $\numPoses$ blocks are of shape $(\dimension \liftedDimension \times
\frac{d(d+1)}{2})$. These blocks are the entries associated with the constraints
$\rotOuterProd = I_d$. The remaining $\numDistMeasures$ blocks are
$(\liftedDimension \times 1)$ and are the entries associated with the
constraints $\lVert \dist_{ij} \rVert_2^2 = 1$.

As the adjoint constraint Jacobian is block-diagonal, it is sufficient to show that each
block has full column rank. This can be done by recognizing that the original
constraints are defining functions of $\St (\liftedDimension, \dimension)$ and
the unit-sphere in $\R^\liftedDimension$ and that each block is the Jacobian
of a single defining function.

By definition, the Jacobian of the defining function of an embedded submanifold
is full rank at any point on the submanifold \cite[Definition
3.10]{boumal2023introduction}. At any feasible point of
\cref{prob:ra-slam-rr-sdp} each low-dimensional variable (e.g., $\rot_i$) is
satisfying its defining function and thus is on its corresponding submanifold.
Therefore, at any feasible point, each block of the adjoint constraint Jacobian
$\ConstraintJacobian$ is full rank. Thus, at any feasible point the adjoint
constraint Jacobian has full column rank and the LICQ is satisfied.
\QED

\comment{
Upon recognizing the underlying Riemannian geometry of the constraints, it is
unsurprising that the LICQ is satisfied for our problem. A reader familiar with
Riemannian geometry may recognize that this proof could be equivalently
recognized as demonstrating that the constraints of the rank-restricted SDP are
a defining function of the product manifold $\St (\liftedDimension,
\dimension)^\numPoses \times \OB (\liftedDimension, \numDistMeasures)$. Indeed,
a similar proof could be made for any rank-restricted SDP which can be
equivalently posed as unconstrained optimization over a Riemannian manifold.
}

\subsection{Performing Certification}

In this section we demonstrate how solution certification is reduced to three
steps: (i) solving a single linear least-squares problem, (ii) matrix addition,
and (iii) evaluating positive semidefiniteness of a real, symmetric matrix. All
matrices in these computations are sparse, enabling efficient computation.

\subsubsection*{Computing Lagrange Multipliers}

As the Lagrange multipliers are a function of the specific solution point
evaluated, they must be computed for each candidate solution. By
the stationarity condition of the KKT conditions, we determine the Lagrange
multipliers via linear least-squares. For any stationary point, the
partial derivative of the Lagrangian with respect to the problem variable is
zero \cite[Theorem 12.1]{nocedal99numerical}. Gathering the Lagrange
multipliers as a vector,
    $\lambda \in \R^{\numConstraints}
    \triangleq
        [\lambda_{1}, \dots , \lambda_{\numConstraints}]$,
the partial derivative of the Lagrangian of \cref{prob:ra-slam-rr-sdp} is
\begin{equation}
    \label{eq:qcqp-stationarity-derivation}
    \begin{aligned}
        \partial_\matX \mathcal{L}(\matX, \lambda)
         & = \partial_\matX \StackedMatrixObjectiveFunction                                            \\
         & \hspace{4mm} + \partial_\matX \sumOverConstraints (\LagrangianConstraintFunction) \lambda_i \\
         & = 2 \matQ \matX + 2 (\sumOverConstraints \lambda_{i} \matAi) \matX.
    \end{aligned}
\end{equation}

It follows from \cref{eq:qcqp-stationarity-derivation} that the first-order
stationarity condition of \cref{prob:ra-slam-rr-sdp} is,
\begin{equation}
    \label{eq:qcqp-stationarity-original}
    \matQ \candidateSolution + (\sumOverConstraints \lambda_{i} \matAi) \candidateSolution = \zeros,
\end{equation}
where $\candidateSolution \in \matXStateSpace$ is a stationary point of
\cref{prob:ra-slam-rr-sdp} and $\zeros$ is the zero matrix.

Observe that \cref{eq:qcqp-stationarity-original} is linear with respect to the
Lagrange multipliers. We rearrange \cref{eq:qcqp-stationarity-original} to
arrive at
\begin{equation}
    \label{eq:qcqp-stationarity-intermediate}
    \sumOverConstraints (\matAi \candidateSolution) \lambda_{i} = - \matQ \candidateSolution.
\end{equation}

By vectorizing each side of \cref{eq:qcqp-stationarity-intermediate} we arrive
at
\begin{equation}
    \label{eq:qcqp-stationarity-llsq}
    \ConstraintJacobian \lambda = - \vect (\matQ \candidateSolution)
\end{equation}
where $\ConstraintJacobian$ is the adjoint constraint Jacobian of
\cref{eq:constraint-jacobian}. Thus, the Lagrange multipliers $\lambda$ may be
estimated by solving the linear system of \cref{eq:qcqp-stationarity-llsq}. As
the adjoint constraint Jacobian is full rank (\cref{thm:licq}), the Lagrange
multipliers are uniquely determined.

\subsubsection*{Building $\certMat$ and Evaluating Optimality}

Given the Lagrange multipliers and constraints, $(\lambda_i, \matAi)$, the
certificate matrix can be formed as in \cref{eq:cert-matrix}:
$\certMat \triangleq Q + \sum_{i=1}^m \lambda_i \matAi$.

By \cite[Theorem 4]{rosen20wafr}, if $\certMat
    \isPSD$, the candidate solution $\candidateSolution$ is globally optimal.
For this problem, the most efficient means of checking $\certMat
    \isPSD$ is by attempting to Cholesky factorize $\certMat + \beta
    I$, where $0 < \beta \ll 1$.
Existence of a factorization ($M = L L^\top$) is sufficient proof of positive
semidefiniteness of $M$ and all positive definite matrices admit a Cholesky
factorization. We can thus interpret $\beta$ as a numerical tolerance parameter
for the positive semidefiniteness of $\certMat$. We summarize this
certification scheme in \cref{alg:certification}.
\begin{algorithm}[!tbp]
    \algblock[TryCatchFinally]{try}{endtry}
    \algcblock[TryCatchFinally]{TryCatchFinally}{finally}{endtry}
    \algcblockdefx[TryCatchFinally]{TryCatchFinally}{catch}{endtry}
        [1]{\textbf{catch} #1}

    \caption{Certify Optimality of a Candidate Solution, $\candidateSolution$}
    \label{alg:certification}
    \textbf{Input:}  the solution to certify $\candidateSolution \in
        \matXStateSpace$, positive semidefinite tolerance parameter $\beta \in \Rpos$
    \\
    \textbf{Output:} whether $\candidateSolution$ is a low-rank
    solution to \cref{prob:ra-slam-convex-sdp}
    \begin{algorithmic}
        \Require $\candidateSolution$ is locally optimal
        \Function{certify}{$\candidateSolution, \beta$}

        \State $\lambda \gets $ solve the linear system of \cref{eq:qcqp-stationarity-llsq}

        \State $\certMat \gets
            \matQ +
            \sum_{\indexVar=1}^\numConstraints \lambda_\indexVar \matAi $

        \try { Cholesky($\certMat + \beta I$)}
        \State{ \Return certified = True}
        \catch{$\certMat + \beta I $ not positive definite}
        \State{ \Return certified = False}
        \endtry
        \EndFunction
    \end{algorithmic}
\end{algorithm}

\section{Certifiably Correct Estimation}
\label{sec:certifiably-correct-estimation}

\CoraAlgorithm

In this section we describe our approach to certifiably correct estimation.
We demonstrate that the rank-restricted SDP of \cref{prob:ra-slam-rr-sdp} can be
solved via Riemannian optimization, enabling use of the Riemannian Staircase
methodology. We then describe the Riemannian Staircase as applied to
\cref{prob:ra-slam-rr-sdp} and discuss how we use the Riemannian Staircase
estimate to extract an estimate to the MAP problem (\cref{prob:ra-slam-map}).


\subsection{The Riemannian Staircase}
\label{sec:riemannian-staircase}

\def\requiredLiftingBoundFootnote{\footnote{\label{footnote:level-of-relaxation}There
        are known bounds on the required $\liftedDimension$ to guarantee such that every
        second-order critical point of \cref{prob:ra-slam-rr-sdp} is globally optimal
        for \cref{prob:ra-slam-convex-sdp} \cite{boumal16neurips,burer05mathprog}. In
        practice, these bounds are often well above the level of relaxation required to
        obtain an optimal solution to \cref{prob:ra-slam-convex-sdp}.}}

The Riemannian Staircase approach we use follows the general methodology
outlined in \cite[Algorithm 1]{boumal15arxiv} with slight alterations.
Specifically, Riemannian optimization is performed at a given level of
relaxation, as determined by $\liftedDimension$. If the estimate
$\candidateSolution$ is found to be optimal (\cref{alg:certification}) then the
algorithm returns the certified solution. If $\candidateSolution$ is not
certified as optimal, $\liftedDimension$ is incremented and the previous
estimate is used to initialize the relaxed problem.  This process continues
until a certifiably optimal solution is
found\requiredLiftingBoundFootnote.

Critically, when incrementing $\liftedDimension$ the estimate will be a
first-order stationary point if lifted by appending a column of zeros to
$\candidateSolution$. We use a saddle escape technique which computes a
second-order descent direction from the negative eigenspace of $\certMat$ to
perturb the solution off the stationary point \cite{rosen20wafr,boumal15arxiv}.

The primary differences between our approach and \cite[Algorithm
    1]{boumal15arxiv} are that the previous work used rank-deficiency of the
estimate to certify the solution. Our certification scheme allows for us to
certify solutions which are not rank-deficient, permitting computational
advantages as we can certify solutions at lower levels of relaxation.
Additionally, we use a specialized eigensolver which allows us to compute the
saddle escape direction typically 1-2 orders of magnitude faster than standard
eigensolvers \cite{rosen22ral}.

\ProjectSolutionAlgorithm

In our implementation of the Riemannian Staircase we use the Riemannian
Trust-Region (RTR) method \cite{absil07fcm}. Inner iterations of the RTR method
solve a trust-region subproblem via a conjugate gradient method. The
computational efficiency of RTR can be highly dependent on the conditioning
of this subproblem.
A key to efficient RTR is to ensure good subproblem conditioning with the
    use of a preconditioner, a linear, symmetric, positive definite operator which
    seeks to approximate the inverse of the \textit{Riemannian Hessian} of the
    objective function.

    Previous works \cite{rosen19ijrr,briales17ral,tian21tro} found the data
    matrix $\matQ$ to closely approximate the Riemannian Hessian. This is
    theoretically justified by the fact that the Riemannian Hessian is closely
    related to the certificate matrix $\certMat = \matQ + \sum \lambda_i \matAi$
    \cite[Appendix B.4.2]{tian21tro}.  As such, the inverse of a regularized
    version of the data matrix $(\matQ + \mu I)^{-1}$ serves as an effective
    preconditioner. This is the approach we take in our implementation, in which
    we cache a Cholesky factorization of $\matQ + \mu I$ at the beginning of the
    optimization and use it to efficiently apply the preconditioner. In
    practice, $\mu$ is chosen to ensure that the
    preconditioner is well-conditioned (typically with a condition number of less
    than $10^6$).

    We verified the efficacy of this preconditioner by comparison to a range of
    other common preconditioners based on $\matQ$, including the Jacobi,
    block-Jacobi, and incomplete Cholesky preconditioners. In all cases, the
    regularized Cholesky preconditioner was found to result in the fastest
    convergence of the RTR method.
    The regularized Cholesky preconditioner and
    other preconditioners are available in our open-source implementation.

\subsection{Certifiably Correct Estimation}

We have now established: a certification scheme, means for lifting the
\cref{prob:ra-slam-rr-sdp} to higher dimensions, and how to represent each
lifted relaxation of \cref{prob:ra-slam-rr-sdp} as a Riemannian optimization
problem. We combine these three contributions to construct an
initialization-independent approach to certifiably correct RA-SLAM (CORA), as
outlined in \cref{alg:estimation}.

Given an initial estimate $\matX_0$ our Riemannian Staircase approach is
applied to obtain a symmetric factor $\candidateSolution$ for a solution of
\cref{prob:ra-slam-convex-sdp}. This SDP solution is projected to the feasible
set of the original problem (\cref{prob:ra-slam-map}) and the projected solution
is refined with local Riemannian optimization.

This refinement is done, as the SDP solution is not guaranteed to be rank
    $d$. If the solution were rank $d$ then (i) the projected solution would
    correspond \textit{exactly} to a minimizer of the original MAP problem
    (\cref{prob:ra-slam-map}) and (ii) the relaxation would be tight. There are
    a number of reasons why the obtained solution may not be rank $d$, including
    practical (e.g., the SDP solution obtained is only an approximate solution
    due to numerical precision) and theoretical (e.g., no rank $d$ solution to
    the SDP exists or the SDP has multiple solutions, only some of which are
    rank $d$). However, even if the projected SDP solution is not a minimizer
    of the MAP problem, the refinement step can result in a solution to the MAP
    problem. In this sense, the SDP is a convex relaxation which is capable of
    exactly solving the MAP problem and, thus, certifying global optimality of
    the MAP solution. When the SDP does not exactly solve the MAP problem, it
    still provides a principled initialization for the MAP problem.

Additionally, by obtaining a certifiably optimal solution to the SDP
(\cref{prob:ra-slam-convex-sdp}) we also establish a lower-bound on the optimal
value of \cref{prob:ra-slam-map}. This lower-bound both provides a means of
certification, and an upper-bound on suboptimality should the SDP relaxation not
be tight.

\section{Experiments}
\label{sec:experiments}

CORA was evaluated through a variety of real-world and simulated RA-SLAM
experiments. Two classes of experiments were performed:
\begin{enumerate}
      \item
            On a series of real-world RA-SLAM problems CORA was compared against
            a state-of-the-art local optimizer typically used in RA-SLAM \cite{gtsam}.
            These tests aimed to understand CORA's ability to obtain
            high-quality solutions on realistic scenarios, as measured by
            rotational and translational errors in the estimated poses.
            These RA-SLAM scenarios span a range of platforms, team sizes,
                  and sensor modalities, as described in
                  \cref{tab:experiments-overview}.
                  Results are summarized in
                  \cref{tab:single-robot-errors,tab:multi-robot-errors,tab:mrclam-errors-2} and
                  \cref{fig:traj-errors-bar-plot}. Visualizations of the estimated
                  trajectories are in Appendix \ref{sec:appendix-trajectories}.
                  (\cref{sec:real-world-evaluation})
      \item
            A series of simulated problems were constructed with parameter
            sweeps of: the number of robots, the number of landmarks, and the
            density of inter-robot range measurements. These tests were
            performed both with and without inter-robot relative pose
            measurements. The aim was to empirically evaluate the the tightness
            of the SDP relaxation (\cref{prob:ra-slam-convex-sdp}) with respect
            to common parameters of RA-SLAM problems.
            (\cref{sec:tightness-of-relaxation})
\end{enumerate}

We have attempted to present a descriptive, yet concise, picture of CORA through
these experiments. However, we recognize that future users may hope to
understand other aspects of CORA or study CORA's performance on a specific
problem class. CORA, the experimental data for all of these
problems, tools to visualize and explore the data, and the tools to generate the
simulated problems are available at \url{\RepoURL}.

\subsection{Real-world Evaluation}
\label{sec:real-world-evaluation}

\ExperimentDescriptionsTable

We compare the trajectories obtained by CORA on a series of real-world RA-SLAM
scenarios. The 3D aerial drone scenario is a new dataset that was collected in
the course of this work and is
described in \cref{sec:drone-experiments}. The
remaining scenarios are taken from the literature \cite{djugash14ijfr,yu23arxiv,leung2011utias}.
We summarize the experiments in \cref{tab:experiments-overview}.

Trajectories are evaluated by comparison to ground truth. The translational and
rotational errors between the estimated and ground truth trajectories are
computed by evo \cite{grupp2017evo}.

For each scenario, we compare the trajectory obtained by CORA to a variety of
baselines. Each baseline uses the GTSAM Levenberg-Marquardt (LM) optimizer
\cite{gtsam} to refine a given initialization. As initialization requirements
differ in the single- and multi-robot cases due to the need to estimate the
relative pose between robots, we consider additional initialization strategies
for the multi-robot case. The strategies we compare against represent a range of
realistic initialization approaches, depending on the information available to
the user. In all cases, CORA was initialized randomly to demonstrate
its initialization-independence.

The \textit{single-robot} initialization strategies are: (i) SCORE, a
state-of-the-art initialization strategy for RA-SLAM \cite{papalia23icra}, (ii)
ground-truth landmark positions (GT), and (iii) random landmark positions (R). In
cases (ii) and (iii), the first poses are taken as ground truth and following
poses are initialized via odometry.

In the \textit{multi-robot} case, we consider the following initialization
strategies: (i) SCORE, (ii) ground-truth starting poses and landmark positions
(GT-GT), (iii) ground-truth starting poses and random landmark positions (GT-R),
(iv) random starting poses and ground-truth landmark positions (R-GT), and (v)
random starting poses and random landmark positions (R-R). In cases (ii) - (v),
for each robot, every pose after the first is initialized via odometry. 
    SCORE was not used for the MR.CLAM experiments, as it did not appropriately
    scale to the large problem size of two of the MR.CLAM sessions.

The corresponding error metrics are shown in
\cref{tab:single-robot-errors,tab:multi-robot-errors,tab:mrclam-errors-2} and plotted in
\cref{fig:traj-errors-bar-plot}. These results are discussed in
\cref{sec:practical-results-discussion}.
We present visualizations of the estimated trajectories in Appendix
\ref{sec:appendix-trajectories} in
\cref{fig:drone-trajectories,fig:tiers-trajectories,fig:plaza-trajectories}.

\DroneExperimentalEquipmentFigure
\SingleRobotExperimentErrorsTable
\TIERSExperimentErrorsTable
\MRCLAMExperimentErrorsTable

\ExperimentalErrorsBarPlotFigure
\subsubsection{Drone Experiments}
\label{sec:drone-experiments}

We present experiments performed ourselves with a hexcopter drone, equipped with six Nooploop
ultra-wideband (UWB) radios and an Intel RealSense D435i camera, and a static
ground station with four Nooploop UWB radios (\cref{fig:drone-equipment}). The UWB sensors were used for
two-way time-of-flight range estimation. Ground truth for the drone and ground
station were obtained from a motion capture system.

Odometry measurements were extracted from camera images via Kimera-VIO
\cite{rosinol20icra}. The drone was flown in two rectangular paths over a
10 meter by 10 meter area at a constant height. A single range measurement
between the drone and ground-station was extracted at 10 Hz
by averaging over all measurements obtained between the drone and
ground-station since the previous measurement. Analysis of this measurement
extraction approach showed that the resulting measurements were accurate
measurements of the range between the center of the drone and the center of the
ground-station.

\RuntimeTable
\subsubsection{Real World Datasets}
\label{sec:real-world-datasets-descriptions}

    In addition to the drone data set collected, we evaluated CORA on an additional
    three real-world datasets, comprising five separate scenarios. These data sets
    were collected by others \cite{djugash14ijfr,yu23arxiv,leung2011utias}; an
    overview of the data sets is provided in \cref{tab:experiments-overview}.

    From the Plaza data set \cite{djugash14ijfr}, we used the Plaza 1 and
    2 scenarios in which an autonomous lawnmower obtained odometry from wheel
    encoders and a gyroscope and range measurements to four UWB sensors placed
    in the environment. The Plaza 1 scenario drove a standard
    `lawnmower' path that balances left and right turns. The Plaza 2 scenario
    consistently turned in the same direction, to demonstrate
    drift in heading estimates that can occur in practice.

    The TIERS data set \cite{yu23arxiv} consists of five robots equipped with
    stereo cameras for visual odometry and UWB sensors to perform inter-robot
    ranging. One robot was stationary, acting as a landmark, while the other
    four robots moved in geometric patterns. The robots never crossed paths.

    From the UTIAS MR.CLAM data set \cite{leung2011utias} we used sessions 2, 4,
    6, and 7. The remaining sessions had substantial drop outs in ground truth data. This data
    set consisted of a team of five robots and fifteen landmarks. All robots and landmarks
    were equipped with barcode fiducials, and robots were equipped with cameras to obtain
    range and bearing measurements to the fiducials. The robots obtained odometry from
    open-loop wheel velocity commands. RA-SLAM scenarios were extracted from these sessions
    by ignoring inter-robot bearing measurements. As a result, the scenarios had
    range and bearing measurements between robots and landmarks, pure range measurements between
    robots, and robot odometry measurements. There was large variety in robot trajectories
    and landmark configurations across the sessions, with many trajectories crossing paths.
\subsubsection{Discussion of Real-World Experiments}
\label{sec:practical-results-discussion}

In all of the experiments CORA obtained a trajectory which was qualitatively
correct despite being initialized with random values. This is supported by the
errors shown in \cref{fig:traj-errors-bar-plot}; in each case CORA obtained
errors which were either lowest, or were comparable to the lowest error
baseline.

In contrast to CORA, the GTSAM baselines demonstrated a strong dependence on
initialization to obtain correct solutions. Notably, in the single-drone
experiment, the baseline which used an initial estimate with the ground truth
landmark position (GT) obtained a clearly suboptimal estimate.
    We believe this is due to the substantial odometry drift of the drone
    experiment, leading to the initialization landing in a poor local minimum.
    This is telling, as even having a high-quality initialization on the
    landmark was insufficient to obtain a high-quality estimate. This hypothesis
    is supported by the fact that the GTSAM baseline which used SCORE (a
    different convex relaxation) to initialize the optimization obtained an
    estimate of comparable quality to CORA whereas both GTSAM baselines which
    used odometry-based initialization obtained dramatically worse estimates.

Furthermore, as expected, the single-robot baselines which used randomly
initialized landmark positions (R) typically obtained low-quality estimates.

On the multi-robot, single-landmark TIERS data set, CORA obtained a trajectory
identical to those estimated with ground truth starting poses (GT-GT and GT-R).
As the TIERS data set has notably high-quality odometry, state estimation with
ground truth initial poses is nearly the same as localizing the single landmark
from a large number of range measurements. As suggested by the large errors
obtained by the baselines which do not use ground truth initial poses (SCORE,
R-GT, and R-R) the problem is substantially more challenging without ground
truth initial poses, as the relative poses between different robots must be
estimated entirely from the range measurements. However, CORA obtained a
high-quality estimate from a random initialization, highlighting (i) its
initialization-independence and (ii) ability to solve challenging RA-SLAM
problems.

    On the multi-robot, multi-landmark MR.CLAM data sets, CORA consistently
    obtained the lowest error estimates. This is particularly notable as the
    MR.CLAM data sets have more robots, landmarks, and poses than the other
    experiments. It is of particular interest that CORA outperformed the GTSAM
    baselines which used ground truth initial poses (GT-GT and GT-R), as these
    baselines are equivalent to a best-case scenario in real world use. This
    speaks to the importance for a method that is robust to initialization, as
    it is likely that the odometry drift in the MR.CLAM data sets led to unideal
    initializations for the GTSAM baselines, causing suboptimal
    estimates.

\subsubsection{Runtime}
\label{sec:runtime}

We compare the runtime of CORA to the GTSAM LM optimizer
\cite{gtsam} on the real-world datasets discussed. Both solvers were run on the same
machine on a single core of an i7-12800H CPU. Each solver was initialized with
the same initial guess, using odometry and random landmark initialization. In
multi-robot experiments the relative alignment between robot trajectories was
also initialized randomly.  For single-robot experiments this corresponds to the
(R) initialization strategy and for multi-robot experiments this corresponds to
the (R-R) initialization strategy, as previously described.

The timing results are shown in \cref{tab:experiment_timing}, with experiments
sorted by increasing number of poses.  As can be seen, in the majority of
experiments (5/8) CORA obtained solutions faster than GTSAM. In two experiments
(Plaza 2 and MR.CLAM 7), CORA obtained solutions in less than half the time of
GTSAM. Additionally, in the three experiments with the largest number of poses,
CORA was consistently over 10 seconds faster than GTSAM.

Of the three experiments where GTSAM was faster, only one (TIERS) demonstrated
GTSAM to be significantly faster, in which CORA took 1.7 times longer than GTSAM.
In the Plaza 1 and MR.CLAM 6 experiments, CORA took 1.04 and 1.25 times longer than
GTSAM, respectively.

An interesting observation is that CORA was consistently faster than GTSAM on
the smallest and largest experiments, while GTSAM was faster on the medium-sized
experiments. This suggests that CORA may be able to uniformly outperform
GTSAM with an improved implementation or a different optimization algorithm for
the Riemannian optimization problem (\cref{prob:ra-slam-riemannian-staircase}).

Importantly, the runtimes presented here are for the entire CORA algorithm
(\cref{alg:estimation}). This algorithm typically involves several rounds of
local optimization until the SDP relaxation is solved, followed by a projection
and refinement step. In contrast, the GTSAM runtime reflects only a single round
of local optimization, which often leads to a suboptimal solution (e.g., see
\cref{fig:tiers-trajectories}). Therefore, it is notable that CORA is
competitive with, and often faster than, GTSAM. This is significant given that
CORA offers numerous theoretical advantages and performs, arguably, more
extensive optimization.

\comment{MR.CLAM Results
data/mrclam/range_only/mrclam2/mrclam2.pyfg 260.274
data/mrclam/range_only/mrclam4/mrclam4.pyfg 213.148
data/mrclam/range_only/mrclam6/mrclam6.pyfg 107.833
data/mrclam/range_only/mrclam7/mrclam7.pyfg 113.074
data/mrclam/range_and_rpm/mrclam2/mrclam2.pyfg 36.9005
data/mrclam/range_and_rpm/mrclam4/mrclam4.pyfg 24.0215
data/mrclam/range_and_rpm/mrclam6/mrclam6.pyfg 12.0347
data/mrclam/range_and_rpm/mrclam7/mrclam7.pyfg 14.1442
}

\ParamSweepSuboptFigure
\subsection{Tightness of the SDP Relaxation}
\begin{table}[!t]
    \centering
    \caption{Parameter sweep values for simulated experiments}
    \label{tab:paramsweep}
    \begin{tabular}{ccc}
        \toprule
        \textbf{Parameter}  & \textbf{Default Value} & \textbf{Sweep Range} \\
        \midrule
        Number of Robots    & 4                      & 2--20                \\
        Number of Landmarks & 2                      & 0--10                \\
        Number of Ranges    & 500                    & 100--2000            \\
        \bottomrule
    \end{tabular}
    \vspace{-5mm}
\end{table}

\label{sec:tightness-of-relaxation}

We perform a series of parameter sweeps in a 2-D simulated environment to
evaluate the tightness of the SDP relaxation (\cref{prob:ra-slam-convex-sdp})
with respect to common parameters of RA-SLAM problems.  As we cannot \textit{in
    general} guarantee that we are obtaining an optimal solution to the MAP problem
(\cref{prob:ra-slam-map}), we attempt to evaluate the tightness of the SDP
relaxation through the optimality gap $f_{cora} - f_{sdp}$, where $f_{cora}$ is
the objective value of the final solution obtained by CORA and $f_{sdp}$ is the
objective value of the SDP solution. This optimality gap is an upper bound on the
gap between the optimal solution to the MAP problem and the SDP solution and thus serves
as a proxy for the tightness of the SDP relaxation. To account for differences in
scale between objective values of different problems, we instead report the relative
optimality gap $\frac{f_{cora} - f_{sdp}}{f_{sdp}}$.

The parameter sweeps are over: the number of robots, the number of landmarks, and
the density of inter-robot range measurements. For each parameter value we
generate 20 random instances.
All measurements to
    landmarks are range measurements, to mimic the most common case of RA-SLAM in
    which landmarks are ranging beacons.

Additionally, we perform parameter sweeps both with and without inter-robot
relative pose measurements (inter-robot loop closures). This allows us to
understand the impact of the underlying pose-graph component on the tightness of
the SDP relaxation. We motivate this study by framing the RA-SLAM problem as the
combination of pose-graph optimization (estimation of nodes connected by
relative pose measurements) and range-based localization (estimation of nodes
connected by range measurements). As the SDP relaxation of pose-graph
optimization is often tight as long as the underlying graph is connected
\cite{rosen19ijrr,tian21tro}, one may wonder whether the SDP relaxation of
RA-SLAM is tight in the case that the underlying pose-graph optimization
component is connected.

We list the values of our parameter sweep in \cref{tab:paramsweep}. For each
parameter sweep the default value is used for all parameters except the one
being swept. Each instance consisted of 4000 robot poses. Experiments with
inter-robot loop closures were performed with 200 inter-robot loop closures.

We show the obtained optimality gaps in \cref{fig:subopt-param-sweeps}, which
demonstrate that in the case of inter-robot loop closures the attainable
relative optimality gap is consistently less than 0.2\%. In the case without
inter-robot loop closures, the relative optimality gap is typically less than
10\%. This still suggests that the SDP relaxation is sufficiently tight to be
useful, but that without inter-robot loop closures the SDP relaxation is not
exactly tight. These results suggest that the connectivity introduced by
inter-robot loop closures plays a substantial role in the tightness of the SDP
relaxation.

\begin{table}[!t]
    \centering
    \caption{Percentage of simulated experiments with zero suboptimality. We
        consider the optimality gap to be zero if the relative optimality gap is
        less than $10^{-5}$.}
    \label{tab:relative-opt-gap}
    \begin{tabularx}{\linewidth}{ccc}
        \toprule
                            & \multicolumn{2}{c}{\textbf{\% Experiments with Zero Suboptimality}}                                  \\
        \cmidrule(lr){2-3}
        \textbf{Parameter}  & \textbf{With Loop Closures}                                         & \textbf{Without Loop Closures} \\
        \midrule
        Number of Robots    & 36.2\%                                                              & 0.7\%                          \\
        Number of Ranges    & 33.3\%                                                              & 0.5\%                          \\
        Number of Landmarks & 31.0\%                                                              & 3.4\%                          \\
        \bottomrule
    \end{tabularx}
    \vspace{-5mm}
\end{table}

Furthermore, in \cref{tab:relative-opt-gap} we show the percentage of experiments with relative
optimality gap less than $10^{-5}$, which we consider to be within numerical
tolerance of zero. As can be seen, many of the experiments with inter-robot loop
closures (over 30\%) had zero relative optimality gap. In contrast, only a
small fraction of the experiments without inter-robot loop closures (on the order of
1\%) had zero relative optimality gap. This suggests that while the proposed relaxation
may be tight both with and without loop closures, the presence of loop closures
can greatly strengthen the tightness of the relaxation.
    These observed relationships are consistent with previous theoretical
    and empirical results that link the tightness of similar SDP relaxations to
    the connectivity of the underlying graph
    \cite{holmes2023efficient,rosen19ijrr}.

This hypothesis on the connection between the tightness of the SDP relaxation
and the connectivity of the underlying pose-graph component is further supported by
the parameter sweeps over the number of robots and the number of landmarks.
With increasing numbers of robots, the relative optimality gap only
increased in the scenario lacking inter-robot loop closures. This is
likely because when lacking inter-robot loop closures, each additional robot
adds unconstrained degrees of freedom to the underlying pose-graph.

Interestingly, the addition of landmarks had differing effects in the cases with
and without inter-robot loop closures. With loop closures, the relative
optimality gap appeared to increase slightly as the number of landmarks
increased. Notably, all such problems with no landmarks were solved to zero
suboptimality. This is likely due to the fact that the no landmark case
with inter-robot loop closures is pose-graph optimization that is
augmented with range measurements. I.e., when considering only relative pose
measurements the measurement graph is a single connected component. The addition
of range measurements serve to further refine
the pose-graph. As our simulation only observed landmarks via range
measurements, the addition of landmarks adds additional degrees of freedom which
must be fully constrained by only range measurements.  This likely leads to
increased symmetry or ambiguity in the solution, which is reflected in a
loosening of the SDP relaxation.

In contrast, the addition of landmarks decreased the relative optimality gap in
the case without inter-robot loop closures. We hypothesize this is because the
landmarks improved the underlying graph connectivity and thus helped to better
constrain the relative poses of the robots.

Addition of range measurements appeared to have no discernable effect in the
case with inter-robot loop closures. This is likely because the loop closures
sufficiently constrain the relative poses of the robots. In contrast, without
loop closures, the addition of range measurements appeared to tighten the
relaxation and result in a decrease in the relative optimality gap. This is
likely because, similar to the case of adding landmarks, the range measurements
improved the graph connectivity such that increasing the number of measurements
better constrained the relative poses of the robots.

We would like to emphasize that while the SDP relaxation is likely not exactly
tight for many RA-SLAM problems, the experiments and results presented in this
section suggest that the SDP relaxation is sufficiently tight to be useful in
practice. We view the observation that the SDP relaxation is not exactly tight
as a point of theoretical interest and motivation for future work, rather than
as a substantial limitation.

\section{Conclusion}
\label{sec:conclusion}

In this work we described CORA, the first algorithm to obtain certifiably
optimal RA-SLAM solutions. CORA combines two novel algorithmic capabilities for
RA-SLAM, (i) a global optimality certification method and (ii) a Riemannian
Staircase methodology for initialization-independent certifiably correct
RA-SLAM.

We demonstrated the efficacy of CORA on a series of real-world experiments.  We
demonstrated that, unlike existing approaches, which often perform well but are
susceptible to poor initialization and local optima, CORA is
initialization-independent and consistently obtains low-error estimates.

Additionally, we studied the tightness of the SDP relaxation which underlies
many of CORA's capabilities. We found that under many realistic conditions the
SDP relaxation is tight (i.e., the relative suboptimality gap is below
$10^{-5}$). Furthermore, in conditions where the SDP relaxation likely is not
tight we still typically obtain relative suboptimality gaps below 10\%,
suggesting that the SDP relaxation is sufficiently tight to be practically
useful.

Our parametric study of SDP tightness pointed to a strong relationship between
the connectivity of the pose-graph component of the RA-SLAM problem and the
tightness of the SDP relaxation, suggesting that a
fully-connected pose-graph component has a strong effect on the tightness of the
SDP relaxation. Additionally, this study suggested that in the absence of a
fully-connected pose-graph component the addition of range measurements and
static landmarks can substantially improve the tightness of the SDP relaxation.
We view further theoretical study of these relationships as promising future
work.
    Other interesting future directions in certifiable RA-SLAM include:
    inclusion of redundant constraints and their effect on the tightness of the SDP
    relaxation (similar to e.g., \cite{yang20tro,brynte2022tightness,dumbgen2023toward}),
    alternative sensor noise models (e.g., \cite{barfoot2023certifiably,yang20tro}),
    and distributed algorithms (e.g., \cite{tian21tro}).

\subsubsection*{Disclaimer}

{
This document does not necessarily reflect the views of any
supporting funding agencies.}
\footnote{\tiny{
    This report was prepared as an account of work sponsored by an agency of the
    United States Government. Neither the United States Government nor any agency
    thereof, nor any of their employees, makes any warranty, express or implied, or
    assumes any legal liability or responsibility for the accuracy, completeness,
    or usefulness of any information, apparatus, product, or process disclosed, or
    represents that its use would not infringe privately owned rights. Reference
    herein to any specific commercial product, process, or service by trade name,
    trademark, manufacturer, or otherwise does not necessarily constitute or imply
    its endorsement, recommendation, or favoring by the United States Government or
    any agency thereof. The views and opinions of authors expressed herein do not
    necessarily state or reflect those of the United States Government or any agency
    thereof.}
}

\begin{appendices}

\DroneTrajFigure
\PlazaTrajFigure

\section{Visualization of Real-World Trajectories}
\label{sec:appendix-trajectories}

In \cref{fig:drone-trajectories,fig:plaza-trajectories,fig:tiers-trajectories}
we provide visualizations of the estimated trajectories from all real-world
experiments except for the MR.CLAM data set, which has trajectories that are
difficult to qualitatively evaluate due to substantial trajectory overlap.  As
can be seen, and is described in the main text, the trajectories estimated by
CORA are highly accurate.

In general, both CORA and GTSAM are capable of obtaining high-quality estimates.
However, there are several cases where GTSAM fails to obtain a high-quality
estimate, despite reasonable initializations. These are the scenarios in which
CORA provides practical advantage, as CORA does not depend on the quality of the
initialization.

\TiersExperimentTrajFigure


\section{Form of the data matrix, $\matQ$}
\label{sec:data-matrix-definition}

\def \matQpose{\matQ_{p}}
\def \matQrange{\matQ_{r}}

$\matQ$ is a sparse, real, symmetric, positive-semidefinite matrix which encodes
the cost of the RA-SLAM problem. We describe $\matQ$ as the summation of the
relative-pose cost terms ($\matQpose$) with the range cost terms ($\matQrange$),
i.e.,
\def\rotDim{\dimension \numPoses}
\def\distMeasureDim{\numDistMeasures}
\def\translationDim{\numPoses}
\begin{align}
    \matQ      & \triangleq \matQpose + \matQrange \\
    \matQpose  & \triangleq
    \begin{bNiceMatrix}
        L(G^{\rho}) + \Sigma & \zerosMat{\rotDim}{\distMeasureDim} & \tilde{V}\\
        \zerosMat{\distMeasureDim}{\rotDim} & \zerosMat{\distMeasureDim}{\distMeasureDim} & \zerosMat{\distMeasureDim}{\translationDim}\\
        \tilde{V}^\top & \zerosMat{\translationDim}{\distMeasureDim} & L(G^\tau)
    \end{bNiceMatrix}                      \\
    \matQrange & \triangleq
    \begin{bNiceMatrix}
        \zerosMat{\rotDim}{\rotDim} & \zerosMat{\rotDim}{\distMeasureDim} & \zerosMat{\rotDim}{\translationDim} \\
        \zerosMat{\distMeasureDim}{\rotDim} & W \tilde{D}^2 & \tilde{D} W C \\
        \zerosMat{\translationDim}{\rotDim} & C^\top W \tilde{D} & C^\top W C \\
    \end{bNiceMatrix},
\end{align}
where the first block column/row is size $\rotDim$ and corresponds to the
rotational variables $\rot_i$, the second block column/row is size
$\distMeasureDim$ and corresponds to the auxiliary distance variables
$\dist_{ij}$, and the last block column/row is size $\translationDim$ and
corresponds to the translational states $\tran_i$.

The block matrices which make up $\matQpose$ are the same matrices defined
in \cite{rosen19ijrr}, which connect the estimation problem to the underlying
graphical structure. Similarly, the block matrices which form $\matQrange$
are identical to the matrices defined in \cite{halsted22arxiv} and also draw
on the graphical structure of the underlying inference problem.

\section{Defining the constraints, $\matAi$}
\label{sec:constraint-matrix-definition}

We define the constraints $\tr (\matAi \matXOuterProd) = b_i$ which arise in the
SDP (\cref{prob:ra-slam-convex-sdp}) and rank-restricted SDP
(\cref{prob:ra-slam-rr-sdp}) problems. Each constraint is defined by a
matrix-scalar pair, $(\matAi, b_i)$, where $\matAi \in \R^{\sizeQcqpStateSpace
        \times \sizeQcqpStateSpace}$ is a real, symmetric matrix and $b_i \in \R$ is a
scalar.

The constraints can be sorted as either orthonormality constraints,
$\rotOuterProd = I_d$, or unit-norm constraints, $\lVert \dist_{ij}\rVert_2^2 =
    1$. As described in \cref{sec:rank-restricted-sdp}, each orthonormality constraint
is $\frac{d(d+1)}{2}$ constraints in the form of $\tr (\matAi \matXOuterProd) =
    b_i$ and each unit-norm constraint is a single constraint in the same form.

Each constraint $\matAi$ is has a block-diagonal sparsity pattern and is nonzero
only in the block entry corresponding to the variable it constrains.

\subsection*{Orthonormality Constraints}

\def \matAjl{\matA_{j, k, l}}
\def \matAjlOline{\overline{\matA}_{j, k, l}}
As they correspond to $\rot_i$ variables, the orthonormality constraints occupy
the upper-left $\numPoses$ $(\dimension \times \dimension)$ block-diagonal
entries of the constraint matrices. As $\rotOuterProd$ is symmetric, it suffices
to explicitly constrain only the upper-triangular entries of $\rotOuterProd$.
The constraint $\matAjl$ denotes the constraint corresponding to the $(k, l)$
entry of $\rot_j^\top \rot_j$. We can write $\matAjl$ as
\begin{equation}
    \matAjl \triangleq \begin{bNiceMatrix}
        0 & \\
        & \Ddots \\
        & & \matAjlOline \\
        & & & \Ddots \\
        & & & & 0 \\
    \end{bNiceMatrix}
\end{equation}
where $\matAjlOline$ is in the $j$th $(\dimension \times \dimension)$
block-diagonal entry.

In the case that $k \neq l$, the $(p, q)$ entry of $\matAjlOline$ is defined as:
\begin{equation}
    \matAjlOline(p, q) \triangleq \begin{cases}
        1/2 & \text{if } (p, q) = (k, l)\\
        1/2 & \text{if } (p, q) = (l, k)\\
        0   & \text{otherwise}
    \end{cases}
\end{equation}
with corresponding scalar $b_{j, k, l} = 0$.

In the case that $k = l$, the $(p, q)$ entry of $\matAjlOline$ is defined as:
\begin{equation}
    \matAjlOline(p, q) \triangleq \begin{cases}
        1 & \text{if } (p, q) = (k, k)\\
        0   & \text{otherwise}
    \end{cases}
\end{equation}
with corresponding scalar $b_{j, k, l} = 1$.

\subsection*{Unit-Norm Constraints}

\def \matAdij{{\matA}_{\dist_{ij}}}
\def \matAdijOline{\overline{\matA}_{\dist_{ij}}}
The unit-norm constraints exist in the $\numDistMeasures$ diagonal entries
following the first $\numPoses$ $(\dimension \times \dimension)$ block-diagonal
entries. We number the $\dist_{ij}$ variables in the order they appear as:
$\dist_1$, $\dist_2$, etcetera. The constraint $\matA_{\dist_l}$ corresponding
the $l$th distance variable, $\dist_l$ is defined as:
\begin{equation}
    \matAdij \triangleq \text{diag}(0_{\dimension \numPoses}, 0_{l-1}, 1,
    0_{\sizeQcqpStateSpace - (l + \dimension \numPoses + 1)})
    \in \R^{\sizeQcqpStateSpace \times \sizeQcqpStateSpace}
\end{equation}
where $0_{v}$ indicates a vector of $v$ zeros, $\text{diag}(\cdot)$ indicates the mapping
from a vector to a diagonal matrix, and $k \triangleq (\dimension + 1) \numPoses
+ \numDistMeasures$ is the height of the matrix $\matX$.


\section{{SDP Relaxation from a QCQP (Shor's Relaxation)}}
\label{sec:shors-relaxation}

    For completeness, we briefly derive the SDP relaxation we use
    (\cref{prob:ra-slam-convex-sdp}) from the QCQP (\cref{prob:ra-slam-qcqp}).
    The relaxation we use is known as Shor's relaxation \cite{shor1987quadratic}
    and is a well-known relaxation for QCQPs.  Our derivation performs a
    variable substitution to convert the QCQP to a (non-convex) SDP, where a
    rank constraint is the source of non-convexity.  To obtain a convex SDP, the
    rank constraint is removed.

We begin by recognizing that the rank $d$ formulation of the
    Burer-Monteiro problem (\cref{prob:ra-slam-rr-sdp}) is exactly equivalent to the
    QCQP problem (\cref{prob:ra-slam-qcqp}). In this case, the variables are
    combined into a single matrix $\matX$ with the block structure
    \begin{equation*}
        \matX \in \R^{k \times d} \triangleq ~~ \stackedMatrixVars.
    \end{equation*}
    
    With the data matrix $\matQ$ defined in Appendix \ref{sec:data-matrix-definition}
    and the constraint pairs $(\matA_i, b_i)$ defined in
    Appendix \ref{sec:constraint-matrix-definition}, it can be readily shown that
    the QCQP can be expressed as
    \begin{equation*}
        \begin{aligned}
             & \min_{\matX \in \R^{k \times d}} ~ \StackedMatrixObjectiveFunction \\
             & \st \StackedMatrixQcqpConstraints .
        \end{aligned}
    \end{equation*}
Now, recognize that the matrix outer product $\matXOuterProd$ implicitly
constructs a positive semidefinite matrix that has rank at most $d$. We can
perform the substitution $\matXOuterProd = \matZ$ with $\matZ \isPSD$ and
$\rank(Z) \leq d$ to arrive at the equivalent non-convex problem
\begin{equation*}
    \begin{aligned}
        \min_{Z \in \QcqpStateSpace} ~ & \SdpObjectiveFunction                                \\
        \st                            & \tr(\matAi Z) = b_i, ~~ i=1, \ldots, \numConstraints \\
                                       & Z \isPSD,                                            \\
                                       & \rank(Z) \leq d.
    \end{aligned}
\end{equation*}

However, the rank constraint is the source of non-convexity in this problem
\cite{vandenberghe1996semidefinite}. To obtain a convex relaxation, we
remove the rank constraint to arrive at the (convex) SDP relaxation
\begin{equation*}
    \begin{aligned}
        \min_{Z \in \QcqpStateSpace} ~ & \SdpObjectiveFunction                                \\
        \st                            & \tr(\matAi Z) = b_i, ~~ i=1, \ldots, \numConstraints \\
                                       & Z \isPSD.
    \end{aligned}
\end{equation*}

Interestingly, there is an alternate common derivation of Shor's relaxation
in which the SDP relaxation is found to be the dual problem of the dual
problem of the original QCQP. In a moment of mathematical elegance, this
process of taking the dual problem twice results in the exact same
relaxation as the derivation we present here.
\\

\section{Landmarks in the Problem Formulation}
\label{sec:landmarks-in-problem-formulation}

We briefly discuss how the formulation of the optimization problems presented
(\cref{prob:ra-slam-map,prob:ra-slam-qcqp,prob:ra-slam-convex-sdp,prob:ra-slam-rr-sdp,prob:ra-slam-riemannian-staircase})
are modified to include landmarks. To simplify the mathematical presentation,
the body of the paper considers only poses as the variables of interest, but the
inclusion of landmarks is straightforward. Importantly, in our implementation,
landmark variables are not represented as `degenerate' poses (pose variables
with rotational components that do not appear in the cost function).  This
detail affects the computational efficiency and numerical stability of the
algorithms we present.  For an investigation of these properties in a problem
formulation that does not consider range measurements, see
\cite{holmes2023efficient}.

We begin by discussing the MAP and QCQP problem formulation
(\cref{prob:ra-slam-map,prob:ra-slam-qcqp}). Additional translational variables
$\tran_i$ need to be introduced for each landmark.  The primary adjustment to
the MAP and QCQP formulations is inclusion of pose-to-landmark measurements,
which often arise in SLAM in the form of e.g., visual feature recognition.
Pose-to-landmark measurements can be included in these problem formulations by
setting the precision of the rotational component $\kappa_{ij}$ of the
measurement to zero.

Inclusion of landmarks in the SDP relaxation and subsequent Burer-Monteiro
factorization (\cref{prob:ra-slam-convex-sdp,prob:ra-slam-rr-sdp}) requires a
different adjustment. Recall that the variables for the QCQP can be combined
into a single matrix $X$ with the block structure
\begin{equation*}
    \matX = \begin{bmatrix}
        {\rot}_1  \dots
        {\rot}_\numPoses \mid  {\dist}_1  \dots
        {\dist}_\numDistMeasures \mid \tran_1 \dots \tran_\numPoses
    \end{bmatrix}^\top.
\end{equation*}
Additionally, recall that for the QCQP, the objective function and constraints can be expressed as the
quadratic forms $\tr{(\matQ \matXOuterProd)}$ and $\tr{(\matA_i \matXOuterProd)}=b_i$ respectively.

From this, a straightforward description of the inclusion of landmarks is by
elimination of rotational variables from the block structure of $\matX$. For example, if
there were $n$ translational states (poses and landmarks) and
$l$ landmarks, the block structure of $\matX$ could be obtained from the previous
form by removing the last $l$ rotational variables. This would result in a
block structure of
\begin{equation*}
    \matX = \begin{bmatrix}
        {\rot}_1  \dots {\rot}_{\numPoses-l} \mid
        {\dist}_1  \dots {\dist}_\numDistMeasures \mid
        \tran_1 \dots \tran_\numPoses
    \end{bmatrix}^\top.
\end{equation*}

Similarly, the rows and columns of $\matQ$ and $\matA_i$ corresponding to the
indices of the deleted rotational variables are removed.

We note that this notation and indexing is convenient for concisely describing
the inclusion of landmarks in the problem formulation. In practice, it is more
convenient to implement the algorithms presented in this paper when landmarks
are counted separately from poses. The most convenient standard we have found
for implementation is, in the case of $\numPoses$ poses and $l$ landmarks, the
following block structure for $\matX$:
\begin{equation*}
    \matX = \begin{bmatrix}
        {\rot}_1  \dots {\rot}_{\numPoses} \mid
        {\dist}_1  \dots {\dist}_\numDistMeasures \mid
        \tran_1 \dots \tran_{\numPoses} \mid
        \tran_{\numPoses+1} \dots \tran_{\numPoses+l}
    \end{bmatrix}^\top,
\end{equation*}
where the last $l$ translational variables correspond to the landmarks.

\end{appendices}

\renewcommand{\bibfont}{\small}
\printbibliography







\newcommand{\vspacestd}{-0mm}

\vspace{\vspacestd}
\begin{IEEEbiography}[{\includegraphics[trim=300 0 300 0, width=1in,height=1.25in,clip,keepaspectratio]
                                {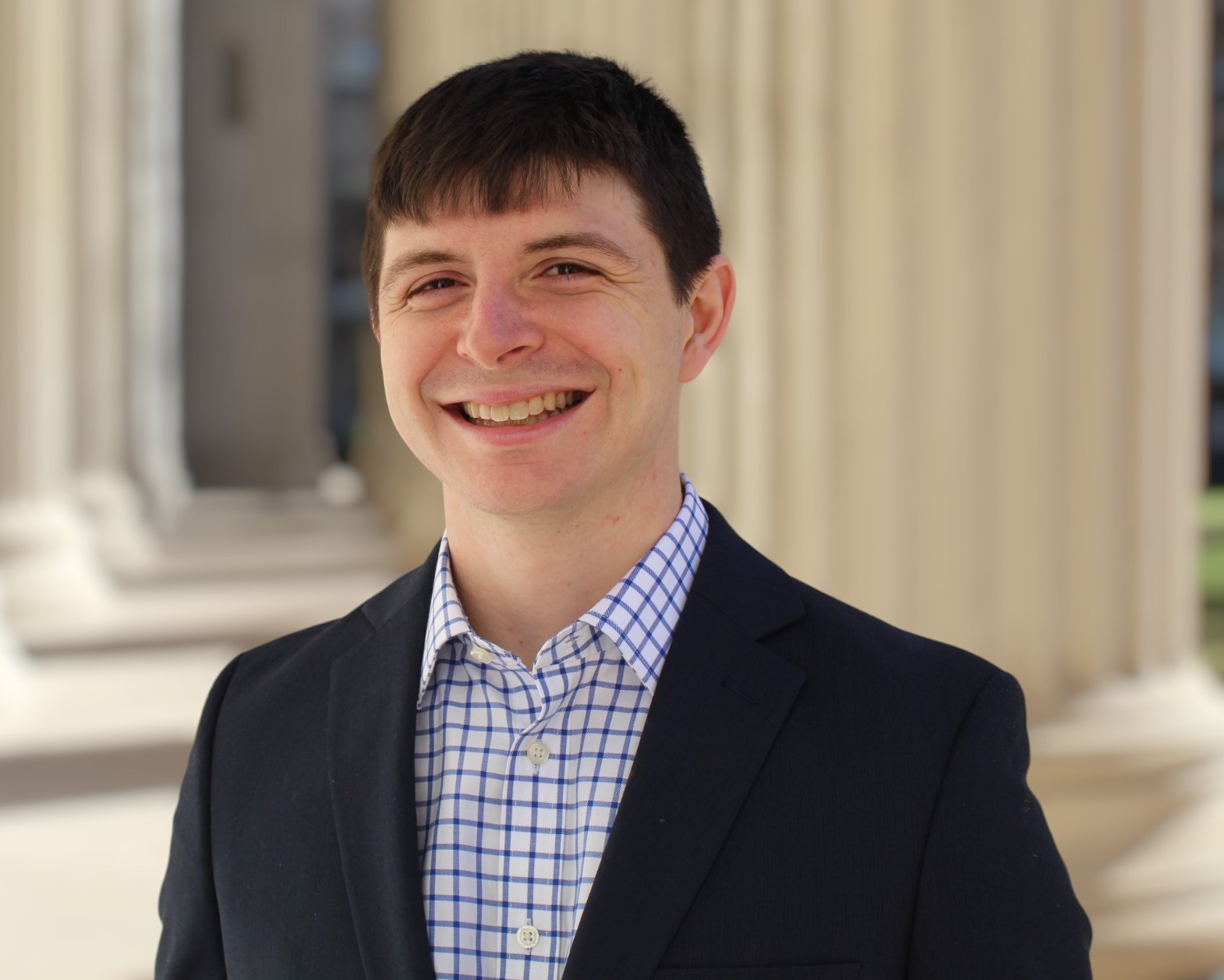}}]{Alan Papalia}
        (Student Member, IEEE)
        received the B.S. degree in mechanical engineering from the
        University of Illinois at Urbana-Champaign (2019). He is
        currently a Ph.D. candidate in the Massachusetts Institute of
        Technology and Woods Hole Oceanographic Institution Joint
        Program in Oceanography/Applied Ocean Science and Engineering.
        His current research interests include applications of
        optimization, graph theory, and information theory to robotic
        navigation, perception, and collaborative autonomy.
\end{IEEEbiography}

\vspace{\vspacestd}
\begin{IEEEbiography}[{\includegraphics[trim=0 15 0 0, width=1in,height=1.25in,clip,keepaspectratio]
                                {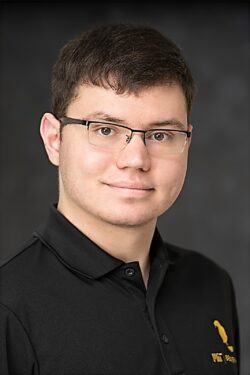}}]{Andrew Fishberg}
        received the B.S. in computer science from Harvey Mudd College
        (2016) before joining MIT Lincoln Laboratory's Advanced Capabilities and
        Systems Group. He then received a S.M. from MIT (2024) in the Department of
        Aeronautics and Astronautics' autonomy and controls tracks, where he is
        currently a Ph.D. candidate. His research interests include optimization,
        range-based localization, and multi-robot systems.
\end{IEEEbiography}

\vspace{\vspacestd}
\begin{IEEEbiography}[{\includegraphics[trim=50 0 50 0, width=1in,height=1.25in,clip,keepaspectratio]
                                {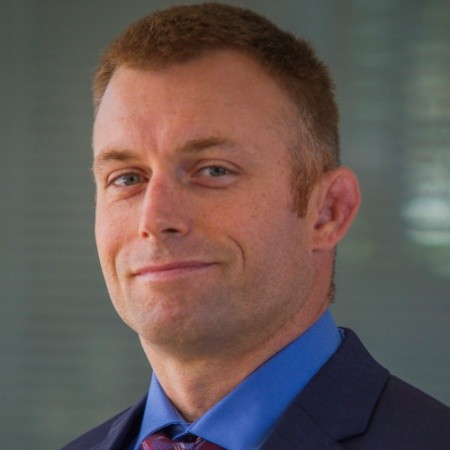}}]{Brendan W. O'Neill}
        received the M.S. degree in mechanical
        engineering from the Massachusetts Institute of Technology (MIT)
        and Woods Hole Oceanographic Institution (WHOI) Joint Program
        (JP) in 2020. He completed his Ph.D. in mechanical and ocean
        engineering in the MIT and WHOI JP in 2023. His current research
        interests include robotic navigation, sensor fusion,
        collaborative autonomy, and human-robot teaming.
\end{IEEEbiography}
\vfill

\vspace{100mm}
\begin{IEEEbiography}[{\includegraphics[trim=430 0 330 0, width=1in,height=1.25in,clip,keepaspectratio]
                                {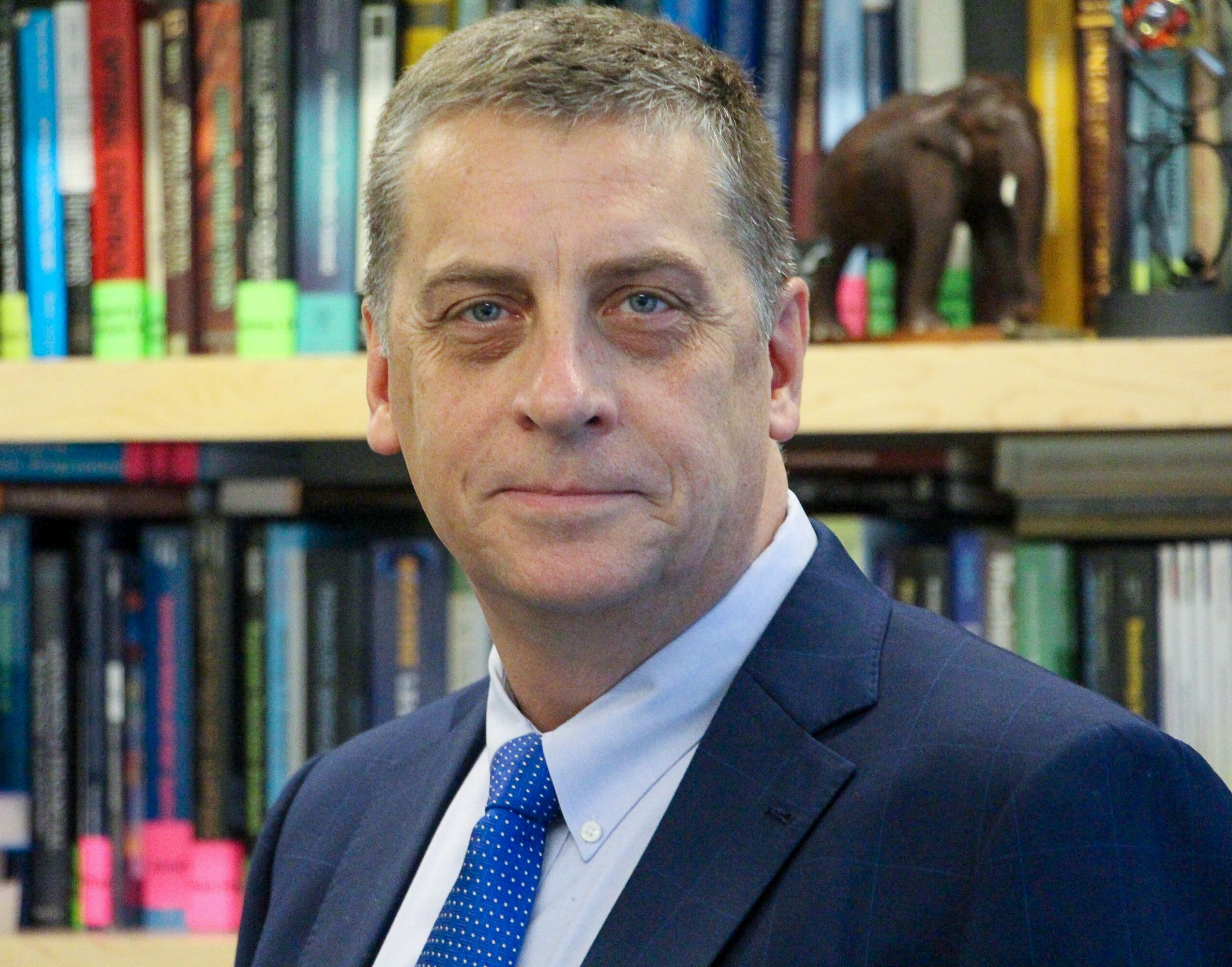}}]{Jonathan P. How}
        (Fellow, IEEE)
        received the B.A.Sc. degree from the University of Toronto, Toronto, ON,
        Canada, in 1987, and the S.M. and Ph.D. degrees in aeronautics and
        astronautics from Massachusetts Institute of Technology (MIT), Cam-
        bridge, MA, USA, in 1990 and 1993, respectively.

        He is currently the
        Richard C. Maclaurin Professor of aeronautics and astronautics with the
        MIT. Prior to joining MIT in 2000, he was an Assistant Professor with
        the Department of Aeronautics and Astronautics, Stanford University.

        Dr. How is a Fellow of AIAA and was elected to the National Academy of
        Engineering in 2021.
\end{IEEEbiography}

\vspace{\vspacestd}
\begin{IEEEbiography}[{\includegraphics[trim=100 0 100 0, width=1in,height=1.25in,clip,keepaspectratio]
                                {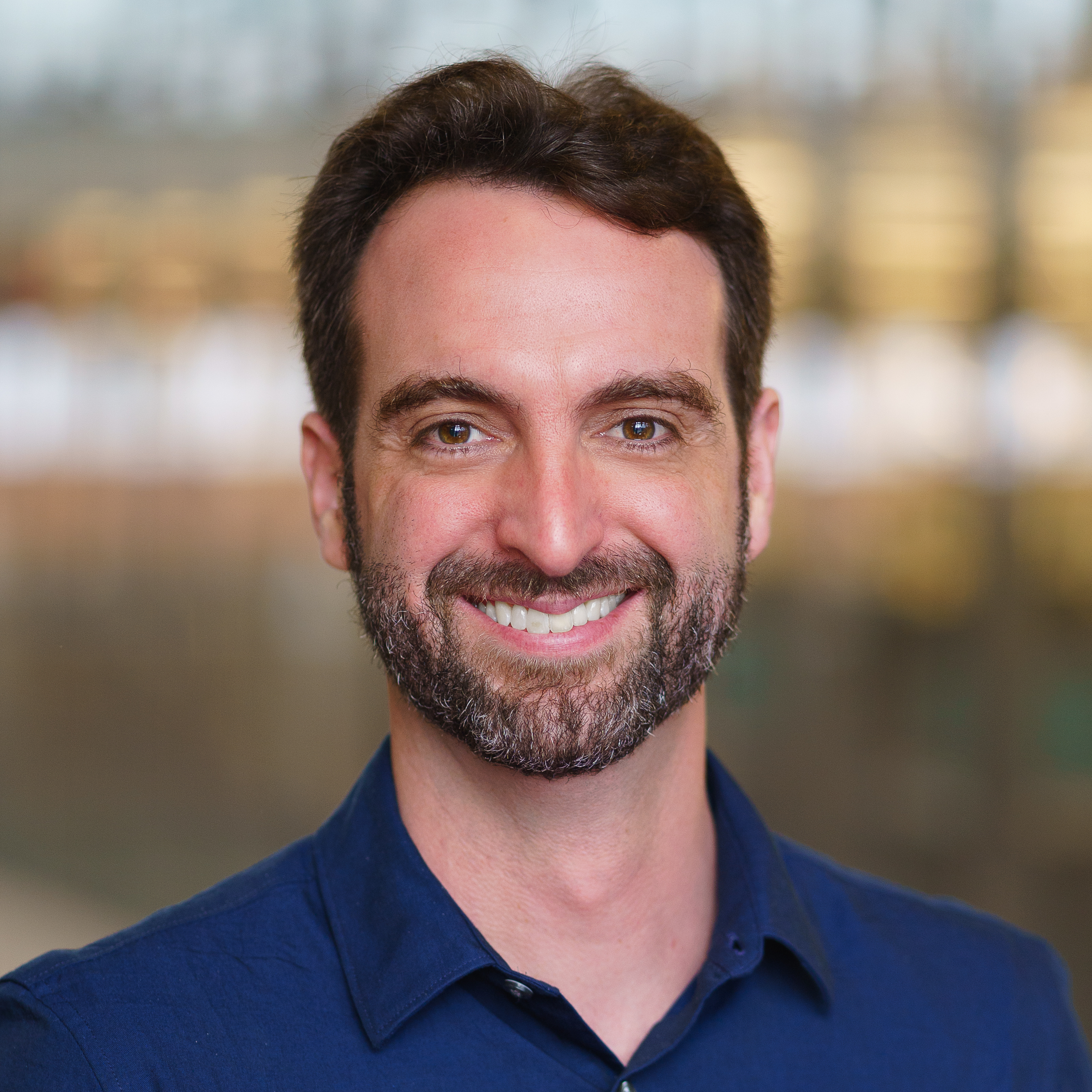}}]{David M. Rosen}
        (Member, IEEE) received the B.S.
        degree in mathematics from the California Institute of
        Technology (2008), the M.A. degree in mathematics from the University of Texas
        at Austin (2010), and the Sc.D.
        degree in computer science from the Massachusetts
        Institute of Technology (2016).

        He is an Assistant Professor with the Departments
        of Electrical \& Computer Engineering and Mathematics, and the Khoury College of Computer Sciences
        (by courtesy) at Northeastern University. His research interest include the
        mathematical and algorithmic foundations of trustworthy autonomy.
\end{IEEEbiography}

\vspace{\vspacestd}
\begin{IEEEbiography}[{\includegraphics[trim=15 0 15 0, width=1in,height=1.25in,clip,keepaspectratio]
                                {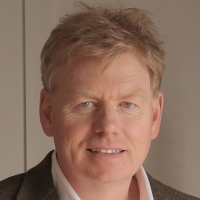}}]{John J. Leonard}
        (Fellow, IEEE) received the
        B.S.E.E. degree in electrical engineering and science from the University of
        Pennsylvania (1987) and the D.Phil. degree in
        Engineering Science from the University of Oxford (1994).

        He is the Samuel C. Collins Professor and Associate Department Head for
        Education in the MIT Department of Mechanical Engineering and a PI in
        the MIT Computer Science and Artificial Intelligence Laboratory (CSAIL).
        He is also a Technical Advisor at Toyota Research Institute. His
        research interests include navigation and mapping for autonomous mobile
        robots.  He is an IEEE Fellow (2014) and an AAAS Fellow (2020).
\end{IEEEbiography}

\vfill

\cleardoublepage


\end{document}